\documentclass[letterpaper, 10 pt, conference]{ieeeconf}  
\usepackage{amsmath}
\usepackage{nicefrac}

\IEEEoverridecommandlockouts                              

\usepackage[pdftex]{graphicx}
\usepackage{import}
\usepackage{xcolor}
\usepackage{colortbl}
\usepackage{units}
\usepackage{caption}
\usepackage{pgf}
\usepackage{tikz}
\usetikzlibrary{arrows,automata}
\usetikzlibrary{positioning}
\tikzset{    state2/.style={ rectangle,  draw=black, inner sep=7pt,  text centered  },}
\usepackage[color=black,opacity=1,angle=0,scale=1]{background}
\backgroundsetup{
contents={
\begin{tikzpicture}
\node at (current page.center) [align=center] {\textcopyright 2022 IEEE. Personal use of this material is permitted. 
\\ Permission from IEEE must be obtained for all other uses, in any current or future media, including reprinting/republishing this material for advertising or promotional purposes, 
\\ creating new collective works, for resale or redistribution to servers or lists, or reuse of any copyrighted component of this work in other works.
\\ DOI: 10.1109/ICRAE56463.2022.10056196};
\end{tikzpicture}},
placement=bottom,
scale=0.6,
vshift=20
}

\title{\LARGE \bf
Importance Filtering with Risk Models for Complex Driving Situations}

\author{Tim Puphal, Raphael Wenzel, Benedict Flade, Malte Probst and Julian Eggert$^*$
\thanks{$^*$ The authors are with the Honda Research Institute Europe GmbH, Carl-Legien-Str. 30, 63073 Offenbach, in Germany. 
\noindent Emails: {\tt\small \{firstname.lastname\}@honda-ri.de}
}}

\begin{document}

\maketitle
\thispagestyle{empty}
\pagestyle{empty}

\begin{abstract}
Self-driving cars face complex driving situations with a large amount of agents when moving in crowded cities. However, some of the agents are actually not influencing the behavior of the self-driving car. Filtering out unimportant agents would inherently simplify the behavior or motion planning task for the system. The planning system can then focus on fewer agents to find optimal behavior solutions for the ego~agent. This is helpful especially in terms of computational efficiency. 
In this paper, therefore, the research topic of importance filtering with~driving risk models is introduced. We give an overview of state-of-the-art risk models and present newly adapted risk models for filtering. Their capability to filter~out surrounding unimportant agents is compared in a large-scale experiment. As it turns out, the novel trajectory distance balances performance, robustness and efficiency well. Based on the results, we can further derive a novel filter architecture with multiple filter steps, for which risk models are recommended for each step, to further improve the robustness. We are confident that this will enable current behavior planning systems to better solve complex situations in everyday driving. 
\vspace*{0.15cm}

\textit{Index Terms}---Driving risk models, object importance, importance filtering, behavior planning, motion planning.

\end{abstract}

\section{Introduction}

Self-driving cars are increasingly able to handle complex driving situations in urban driving environments. Prominent behavior planning systems in research are introduced, e.g., in \cite{ferguson2008, urtasun2019} and \cite{bouton2019}. These systems have sophisticated planning algorithms to tackle complex situations. For example, cooperative planners~\cite{wenzel2021} provide multiple predictions of the future behavior for sensed agents and plan for each predicted scenario an optimal behavior for the ego agent. That means, the system does not only predict a constant velocity trajectory but also anticipates braking and accelerating trajectories~for the other agents. Cooperative planners can thus tackle complex and ambiguous driving situations. At the same time, however, these planners become computationally costly and for situations with many agents, extensive computations are hardly feasible for real-time applications.

In this paper, we therefore consider the task of importance filtering in order to simplify the motion planning problem beforehand. One solution to the aforementioned problem of high computation time is to filter out agents based on their importance, since, by definition, ``unimportant'' agents do not influence the own behavior of the self-driving car. Thus, such agents do not need to be considered in the behavior planner. The core research questions this paper targets to raise and answer are:
\begin{itemize}
 \item Which surrounding agents influence the own behavior and are important?
 \item Can some unimportant agents be neglected before the planning step without inducing safety issues?
 \item After the filtering, is the importance for agents in the remaining group still meaningfully different?
\end{itemize}
Prerequisites we assume given for importance filtering are a sensor setup detecting all surrounding agents, and future paths for the agents from map data or other means. 

\begin{figure}[h!]
  \centering
  \vspace*{0.3cm}
  \resizebox{0.88\linewidth}{!}{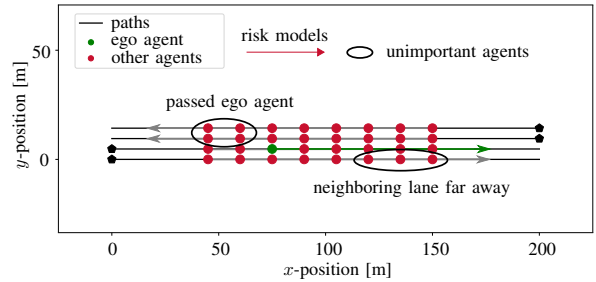}
  \vspace*{0.2cm}
  \caption[]{\hspace{0.07cm} The image shows a complex driving situation with many surrounding agents. By filtering out unimportant agents with driving risk models, the motion planning task can be simplified. In this paper, we compare different risk models on the task of importance filtering.} 
  \label{fig:filtering_toy_example}
\end{figure}

The filtering task is described in more detail in Fig. \ref{fig:filtering_toy_example}, which depicts a complex driving situation with one ego agent (green) and many surrounding other agents (red). Filtering implies to first find out for each other agent what importance it has for the ego agent. We propose to solve this problem by using risk models. The risk the agent induces for the ego agent is computed by using the current state and the paths of the agents. In this way, by setting a threshold that defines for which risk the ego agent is not influenced, some agents can be filtered out to reduce the overall complexity of the problem. Hereby, it is important to not induce safety issues, which we achieve by using risk models. In the toy example of the figure, the filtered out agents are encircled. In this example, these are agents which have passed the ego agent or agents which are on a neighboring lane far away.

\begin{figure*}[!ht]
  \centering
  \vspace{0.1cm}
  \resizebox{0.9\linewidth}{!}{
\begingroup%
  \makeatletter%
  \providecommand\color[2][]{%
    \errmessage{(Inkscape) Color is used for the text in Inkscape, but the package 'color.sty' is not loaded}%
    \renewcommand\color[2][]{}%
  }%
  \providecommand\transparent[1]{%
    \errmessage{(Inkscape) Transparency is used (non-zero) for the text in Inkscape, but the package 'transparent.sty' is not loaded}%
    \renewcommand\transparent[1]{}%
  }%
  \providecommand\rotatebox[2]{#2}%
  \newcommand*\fsize{\dimexpr\f@size pt\relax}%
  \newcommand*\lineheight[1]{\fontsize{\fsize}{#1\fsize}\selectfont}%
  \ifx\svgwidth\undefined%
    \setlength{\unitlength}{681.4710083bp}%
    \ifx\svgscale\undefined%
      \relax%
    \else%
      \setlength{\unitlength}{\unitlength * \real{\svgscale}}%
    \fi%
  \else%
    \setlength{\unitlength}{\svgwidth}%
  \fi%
  \global\let\svgwidth\undefined%
  \global\let\svgscale\undefined%
  \makeatother%
  \begin{picture}(1,0.20416249)%
    \lineheight{1}%
    \setlength\tabcolsep{0pt}%
    \put(0,0){\includegraphics[width=\unitlength,page=1]{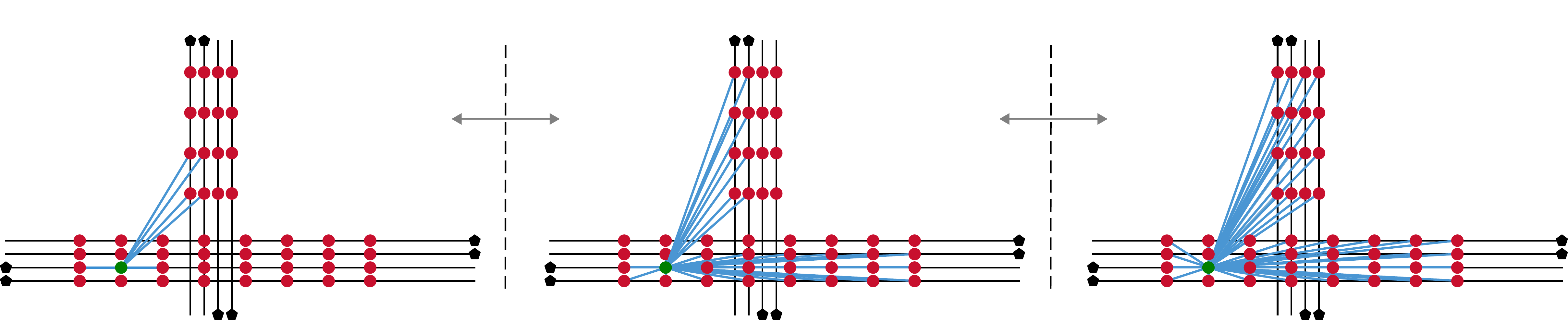}}%
    \put(0.07902888,0.19870233){\color[rgb]{0,0,0}\makebox(0,0)[lt]{\lineheight{1.25}\smash{\begin{tabular}[t]{l}\textbf{high} risk threshold\end{tabular}}}}%
    \put(0.01954952,0.09752607){\color[rgb]{0,0,0}\makebox(0,0)[lt]{\lineheight{1.25}\smash{\begin{tabular}[t]{l}many agents\\filtered out\end{tabular}}}}%
    \put(0.41805397,0.19638646){\color[rgb]{0,0,0}\makebox(0,0)[lt]{\lineheight{1.25}\smash{\begin{tabular}[t]{l}\textbf{medium} risk threshold\end{tabular}}}}%
    \put(0.35793865,0.09752607){\color[rgb]{0,0,0}\makebox(0,0)[lt]{\lineheight{1.25}\smash{\begin{tabular}[t]{l}some agents \\filtered out\end{tabular}}}}%
    \put(0.77363835,0.19804337){\color[rgb]{0,0,0}\makebox(0,0)[lt]{\lineheight{1.25}\smash{\begin{tabular}[t]{l}\textbf{low} risk threshold\end{tabular}}}}%
    \put(0.70338098,0.09752607){\color[rgb]{0,0,0}\makebox(0,0)[lt]{\lineheight{1.25}\smash{\begin{tabular}[t]{l}no agents\\filtered out\end{tabular}}}}%
  \end{picture}%
\endgroup%
}
  \vspace*{0.2cm}
  
  \caption[]{\hspace{0.07cm} We consider importance filtering as a thresholding process for risk values. If the risk value for an agent is smaller than a threshold, the agent is filtered out. In this figure, we show different thresholds for the risk values and the remaining agents after the filtering, which are connected to the ego agent (green) by a blue line. Filtering out agents is essentially equivalent to finding the right threshold.} 
  \label{fig:filtering_and_thresholds}
\end{figure*}
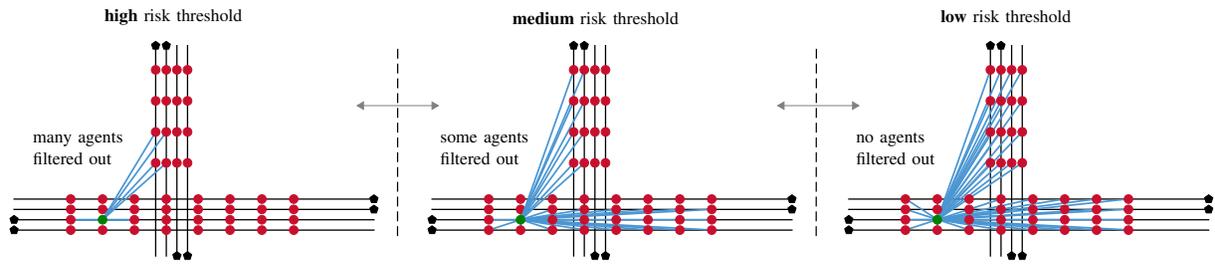

\subsection{Related Work}
When looking at related work for the topic of importance filtering, to the knowledge of the authors, there is few related work. Note that filtering does not refer to tracking a sensed agent over time and reducing inherent noise but neglecting single agents completely to reduce the situation complexity for behavior or motion planners.

Most works develop heuristics to filter out agents, e.g., if there are parked cars behind a barrier, they can be neglected. For dynamic agents, filtering is more complicated and must be achieved more intelligently. In this context, there is only limited research about a higher-level importance measure regarding dynamic agents. 

For instance, the work of \cite{wheeler2016} handles agents' importance based on semantic relations. The authors construct a factor graph that represents situation distributions and in the distributions, they specify inter-vehicle relations and their relation strength, such as a following relation or a neighboring lane relation. Another work \cite{malla2020} develops an importance measure based on vision. They use video data, detect dynamic objects and make trajectory predictions based on such data with machine learning. A degree of influence to the ego agent can now be inferred. Finally, the work of \cite{sugiura2007} develops a planner that derives importance based on fixed discrete states. Sensed objects are classified into  spatial zones (green, yellow and red) and for each zone, a different planning strategy is then activated. As an example, in the yellow zone a virtual force is applied to the robot to keep a sufficient distance between the robot and a classified object. 

Related work on risk models are compared to related work on importance filtering extensive and diverse. Driving risk models, such as from the works of \cite{damerow2014, wachenfeld2016, highway2010, puphal2019, eggert2014}, are mainly applied for the validation of behaviors, motions and maneuvers from self-driving cars, or are applied in a cost function, which is used by the planner itself. Nevertheless, to the knowledge of the authors, a systematic way of using risk models for filtering out surrounding agents based on importance has not been proposed so far.

\subsection{Contribution}

In this paper, we want to build upon the mentioned related work and connect the research topic of importance filtering with driving risk models. We consider the contributions of this paper as three-fold and these are described one by one throughout the paper:
\begin{enumerate}
\item We categorize state-of-the-art risk models and propose newly adapted risk models for filtering. 
\item The models are compared in a large-scale experiment. Hereby, the novel trajectory distance is shown to balance performance, robustness and efficiency well.
\item Based on the results, we furthermore derive a novel filter architecture with multiple filter steps, in which risk models are recommended for each step.
\end{enumerate}
The filtering itself with risk models helps to improve the computational efficiency and the filtering architecture helps to further improve robustness of a behavior planning system. We are confident that this will allow current systems to solve complex and crowded situations with many other agents in everyday driving.

In the next Section~\ref{sec:filtering_and_risk}, the filtering task and its connection to risk are explained in more detail. This section gives more examples of agents that are filtered out depending on a risk threshold and shows inherent challenges of filtering. The following Section \ref{sec:overview_risk_models} then gives an overview of state-of-the-art risk models and presents the newly adapted risk models for filtering. In Section \ref{sec:experiments}, these risk models are then applied on a dataset of over five hundred driving situations with many agents. 
Section \ref{sec:filtering_architecture} outlines the novel filtering architecture, which is derived from the results, and Section \ref{sec:conclusion} finally outlines a conclusion and outlook. 


\section{Importance Filtering and Risk}
\label{sec:filtering_and_risk}
As already mentioned, the task of importance filtering is to filter out unimportant surrounding agents. By computing an importance based on risk models, we consider unimportant agents as agents that the ego agent will most probably not collide with in the future. These agents can therefore be neglected without introducing safety issues. In contrast, we define important agents as agents the ego agent could collide with. Naturally, these agents need to be considered by the behavior planner. 

Importance filtering with risk consists hereby of two steps. In a first step, the contribution of risk for each single agent to the complete situation risk for the ego agent is estimated. Then, filtering implies to choose and set a threshold of one risk value in a second step. If the risk value of one agent is smaller than the threshold, the agent is filtered out. If not, the agent is considered. However, filtering out the right set of agents highly depends on setting the right threshold. 

To understand the challenges of this filtering task, Fig. \ref{fig:filtering_and_thresholds} shows three examples of different filtering thresholds. The agents that are connected to the ego agent (green) with a blue line are found to be important and considered by the planner. All other agents are filtered out. On the left, filtering with a high threshold is shown. Many agents are considered unimportant and are thus filtered out. Hereby, it can happen that agents, which are filtered out, are actually important. We consider these error cases as False Negatives (FN). With a high risk threshold, a large number of FN may happen. In the middle, a medium risk threshold is chosen. Not many agents are filtered out and more are considered as important. In this case, only some agents might induce FN. However, there may be agents in the set that are actually unimportant. Unimportant agents, which are not filtered out, are defined as errors from False Positives (FP). This case also has a small amount of FP. Finally, on the right, an example with a low risk threshold is shown. Not filtering out any agent does not lead to FN but to a large number of errors from FP.

The examples show that it is crucial to correctly calibrate the threshold for the risk models. In the end, the importance filter should only filter out real unimportant agents without causing errors due to FN, while still passing on the important agents without including errors due to FP. In reality, this is not possible and for filtering, we consider FN as more critical than FP. Errors from FN might induce safety issues by the planner by not reacting to an agent which we might collide with. Errors from FP only increases the computation cost for the behavior planner, which reduces the purpose of the filter but does not impair the system functionality. 

The examples also indicate that the performance of the risk model is a determining factor for the threshold setting. We define performance similar to risk accuracy. An accurate model allows to choose the threshold high and filter out many agents, while still having no FN and a low number of FP. An inaccurate model that cannot distinguish between important and unimportant agents will have a worse FN and FP balance so that a low risk threshold has to be chosen. 

We want to finally combine these insights into an easier formulation in order to use them in the experiments section. To achieve this, we define the common terms in classification tasks, which are the True Positive Rate (TPR) and the False Positive Rate (FPR)
\begin{align}
\text{TPR} &= \frac{\text{TP}}{\text{TP}+\text{FN}}, \\ 
\text{FPR} &= \frac{\text{FP}}{\text{FP}+\text{TN}}.
\end{align}
Hereby, these rates weight the correct classifications with the errors from FN and FP. The indicator TPR therein contains the number of correctly classified important agents, called True Positives (TP), and the indicator FPR contains the number of correctly filtered out agents, called True Negatives (TN). In summary, to achieve no errors from FN, a risk model should accordingly reach a value of one for the TPR. At the same time to have only some errors in FP, a value close to zero should be obtained for the FPR. 

\section{Overview of Risk Models}
\label{sec:overview_risk_models}
After explaining the challenges of importance filtering and the relationship with risk, in the following, an overview of the state-of-the-art risk models in the field of automated driving is given. We will classify these models and present newly adapted risk models for filtering. Note hereby that risk is the product of collision probability and severity of the outcome, such as the damage costs for the collision. In this paper, we only focus on probability of a collision. The reason is that the influence of the probability is stronger for planning than the severity. For the filtering task, it is therefore sufficient to concentrate on the collision probability. The severity can be added afterwards in the behavior planner. 

Risk models calculating collision probability can be classified into three classes: distance models, time models and stochastic models. Tab. \ref{tab:overview_risk_models} shows a table with all models for each class that are considered in this paper. The simpler models are in the upper rows and the more complex models in the lower rows. Complex models have a better performance but also higher computational cost. Note that the models which are newly proposed are marked in italics.
%
\subsection{Distance Models}

The simplest model for risk is the current distance. This first distance model only uses the distance $d(t)$ between the ego and another agent at the current time $t$ as an input. If the positions of the agents are defined in two-dimensional space with $\mathbf{x}_1 = [x_1, y_1]$ and $\mathbf{x}_2=[x_2, y_2]$, the current distance follows as
\begin{equation}
d(t) = ||\mathbf{x}_2(t) - \mathbf{x}_1(t)||,
\end{equation}
whereby the risk is its inverse and we write
\begin{equation}
 R_d(t) = \frac{\epsilon}{\epsilon+d(t)}.
\end{equation}

\noindent Generally, the relationship is that the lower the distance the higher the risk. Adding a constant $\epsilon$ allows to normalize the highest risk value to one. Hereby, the agents' shape can be further added in the distance calculation to improve accuracy. An advantage of using this model as a risk model is that it needs a single operation with a time complexity of $\text{O}(1)$ and is thus computationally cheap. However, as we will see, the performance of the model is low.

The second distance model which is commonly used is the path distance. It takes additionally into account the map paths $\mathbf{p}_i$ of the agents as an approximation for their future motion. One path is parametrized by a parameter $m$, which defines the current considered point in the path. A value of $m=0$ could, for example, be the current agent position $\textbf{x}_i$ and $m~=~\infty$ the end of the path. 
The path distance is the minimum distance over all point combinations between both parametrized paths. Accordingly, we define the risk as
\begin{equation}
 R_p(t) = \dfrac{\epsilon}{\epsilon+\text{min}_{m_1,m_2} \{ \hspace{0.03cm} || \mathbf{p}_2(m_2) - \mathbf{p}_1(m_1)|| \} },
\end{equation}
with a constant $\epsilon$. If the paths cross at one~point, the path distance becomes zero and the risk is reaching its maximal value of one. Compared to the current distance, adding a notion of future motion by considering the agents' paths improves accuracy. Since one path contains $p$ points and the distances are computed over each point combination between two paths, the time complexity is given by $\text{O}(p^2)$. 


Both the current distance and the path distance are computationally cheap in their implementation. Nevertheless, a drawback of the two models is that they do not include the velocities of the agents and do not make a detailed prediction of the motion. This can lead to significant errors, especially, when considering a pair of agents driving with complex and dynamic behavioral changes.  

\begin{table}[t!]
  \centering
  \normalsize
  \vspace{0.34cm}
  \resizebox{\columnwidth}{!}{
  \begin{tabular}{|p{3.0cm}|p{4.3cm}|p{3.3cm}|} 
  \hline
    \textbf{Distance models} & \textbf{Time models} & \textbf{Stochastic models} \\
  \hline
    Current distance & Closest encounter \cite{damerow2014} & \textit{Circle approximation} \\ 
  \hline
    Path distance & \textit{Closest encounter and head- way} & 2D Gaussians \cite{puphal2019} \\
  \hline
     \textit{Trajectory distance} & \textit{Closest encounter and 2D headway} & Survival analysis \cite{eggert2014} \\ 
  \hline
  \end{tabular}
  }
  \vspace{0.1cm}
  \caption[]{\hspace{0.07cm} The table gives an overview of the considered risk models. They can be divided into distance models, time models and stochastic models. We mark novel models which are proposed in this paper in italics. These are the trajectory distance, the closest encounter and headway, the closest encounter and 2D headway and the circle approximation.}  
  \label{tab:overview_risk_models}
  \vspace{-0.1cm}
\end{table}

\subsection{Time Models}
In order to include the agents' velocity when calculating risk, time models are often used. The first risk model considered in this paper in the class of time models is the closest encounter model \cite{damerow2014}. In its simplest form, closest encounter assumes constant velocity for the agents. Concretely, the model initially predicts the agents' positions along their path based on their current velocity. In this way, a trajectory $\mathbf{x}_i(t+s)$ is retrieved which depends on the prediction time $t+s$. Please note that the index $i$ denotes the agent number. The model afterwards takes the minimum of all distances at each predicted timesteps which is given with the formula $d(t+s)=||\mathbf{x}_2(t+s)~-~\mathbf{x}_1(t+s)||$. The distance and time of closest encounter can be finally written as
\begin{equation}
 d_E(t) = \text{min}_s\{d(t+s)\}\text{ and } s_E(t) = \text{argmin}_s\{d(t+s)\}.
\end{equation}
If the time of closest encounter is smaller than a maximum value, which means $s_E(t) < s_\text{max}$ holds, the risk for the current time $t$ is now computed with
\begin{equation}
\label{eq:risk_encounter}
 R_E(t) =  \frac{\epsilon}{\epsilon+d_E(t)}.
\end{equation}
The time complexity of the closest encounter is dependent on the number of predicted timesteps $n$, which are considered in the trajectories of the agents and which leads to $\text{O}(n)$. 
A drawback of the closest encounter is that another agent can always deviate from the initially assumed constant velocity behavior. This can lead to risk inaccuracies, especially for cases where the other agent brakes strongly.

To consider behavior uncertainty and the possibility of the other agent braking, the headway \cite{highway2010} can be used as a heuristic. Headway describes the time until the ego agent is reaching the current position of the other agent. This is similar to making a prediction that the other agent suddenly comes to a full stop. When we formulate the headway in one dimension, we can get 
\begin{equation}
\text{TH} = \frac{\Delta l}{v_1} = \frac{l_2 - l_1}{v_1},
\end{equation}
with the longitudinal distance of the agents along the path $\Delta l$ and the velocity of the ego agent $v_1$. The inverse of TH correlates with the risk value $R_{\text{TH}}(t)$ and by setting risk thresholds for both the closest encounter and the headway, we can combine these time models for filtering.

Another drawback is still the restraint that the headway only works for agents on the same path since it is formulated with the longitudinal distance. However, agents on~another path crossing the ego path may also pose a braking risk. In this circumstance, the headway can be included by projecting the other agent from its own path $\mathbf{p}_2$ to the ego path $\mathbf{p}_1$. In this process, the distance to the crossing point is taken as a reference to project the other agent's position with the same distance to the ego path. By calculating the headway again longitudinally, we achieve a two-dimensional form of the headway and can formulate a time-based risk model $R_{\text{2D-TH}}$. The closest encounter, a novel combination of closest encounter and headway, as well as a novel combination of closest encounter and the two-dimensional (2D) headway are listed in Tab. \ref{tab:overview_risk_models}.

\subsection{Stochastic Models}
The last class of risk models that are relevant in this paper and will be described in the following are stochastic models. Instead of considering discrete predictions of constant velocity or braking, stochastic models make use of probability distributions to account for uncertainties in the behavior. This has the advantage that the risk is based on physical stochastic properties. The assumption for uncertainties is that they grow over future time. This can be, e.g., explained by the growing behavior possibilities that other agents possess over time. Furthermore, the sensor uncertainty can be modeled by the initial assumed uncertainty at the current time of the behavior prediction.

The first stochastic model we propose is the novel circle approximation which considers uncertainty by taking a radial distribution of the future positions and approximating the distribution by geometric circles. The circles have a center $\textbf{c}_i$ and a radius $r_i$ for each agent $i$. At each predicted timestep, the minimal distance is computed between the ego circle and the circles of the other agent $j$. In the end, the minimum is also taken over all predicted timesteps $s$. The circle distance is therefore defined as 
\begin{equation}
d_c(t) = \text{min}_j \text{min}_s \textbf{c}_{2,j}(t+s) - \textbf{c}_{1}(t+s) - (r_1(t+s) + r_2(t+s)).
\end{equation}
whereby the risk $R_c(t)$ can be obtained as the inverse, similar to the time models.
Regarding the time complexity, the circle approximation considers $n$ timesteps for the trajectories and additionally formulates $k$ circles for another agent. The time complexity results to $\text{O}(k\cdot n)$. 

While the circle approximation approximates a probability distribution, the second stochastic risk model, the 2D Gaussian method \cite{puphal2019}, formulates probability distributions that are defined by mean position values $\boldsymbol{\mu}_i$ and uncertainties $\boldsymbol{\Sigma}_i$ in $x$- and $y$-direction for the agents. The notion is the same as in the circle approximation, the difference is that an overlap area is calculated between two probability distributions for each predicted time. The overlap between two Gaussians is another Gaussian distribution, which we call $P_o$ and which simplifies the computation. For the risk, we obtain the formula of
\begin{align}
R_G(t) = \text{max}_{s} \operatorname{det}|2\pi&(\mathbf{\Sigma}_{1}+\mathbf{\Sigma}_{2})|^{-\frac{1}{2}} * \nonumber \\ \exp\{-\frac{1}{2}
(\boldsymbol{\mu}_{2}-\boldsymbol{\mu}_{1})^T(\mathbf{\Sigma}_{1}&+\mathbf{\Sigma}_{2})^{-1}
(\boldsymbol{\mu}_{2}-\boldsymbol{\mu}_{1})\}.   
\label{eq:prodgauss}
\end{align}
In this equation, $\mathbf{\Sigma}_{1,2}$ and $\boldsymbol{\mu}_{1,2}$ are functions of the predicted time $t+s$ and the maximum area overlap is used as the risk measure $R_G(t)$ for the current time $t$. In simple form, we can thus write $R_G(t) = \text{max}_s P_o(t+s)$ as the risk value. The 2D Gaussian method has a time complexity similar to the circle approximation but is assumed to be more accurate since it considers real distributions. 

The final stochastic risk model we consider to be most important is the survival analysis \cite{eggert2014}. The survival analysis uses the Gaussian method and combines the method with a Poisson process. Instead of taking the maximum value over all times, the Poisson process allows to integrate the overlaps $P_o(t+s)$ over the prediction steps $s$ and retrieve a risk value. This is achieved by a survival function $S(s;t,\textbf{z}_{t:t+s})$ that is dependent on the current time $t$ and the  future evolution $\textbf{z}_{t:t+s}$ of the driving situation, which is determined by the combination of the agents' trajectories. When defining a critical event rate as the collision overlap over an infinitesimal timestep $\Delta t$ with $\tau_{\text{crit}}^{-1}(\textbf{z}_{t:t+s}) = P_o(t+s)/\Delta t$, we retrieve the risk as
\begin{equation}\label{IntRisk}
R_S(t) = \int_0^{\infty} \tau_{\text{crit}}^{-1}(\textbf{z}_{t:t+s})S(s;t,\textbf{z}_{t:t+s})\,ds.
\end{equation} 

The Poisson process not only has the advantage to consider the collision probability in a fine-grained way by  integrating the probabilities, but it also allows to consider further critical events besides collision events. As the method was already presented previously, we do not describe the survival analysis in detail and refer to the work of \cite{eggert2014}. In summary, the time complexity can be derived to $\text{O}(k\cdot n^2)$ because of the integration over time. Hereby, for simplicity, we added a similar factor $k$ as in the circle approximation. 

\subsection{Novel Trajectory Distance}

Until now, state-of-the-art risk models were shown, which could be classified into distance models, time models and stochastic models. Additionally, three newly adapted models were presented: closest encounter and headway, closest encounter and 2D headway, and the circle approximation. All of the presented risk models have either low performance and low computation time, or high performance and high computation time. We therefore developed a novel distance model, called dynamic horizon trajectory distance, that balances performance, robustness and computation time well for the special task of filtering. 

The trajectory distance combines advantages of distance models (computation time) with those of time models (inclusion of dynamics) to achieve an accuracy close to stochastic models. Concretely, the model includes the velocity $v_i$ from the time models by computing the total traveled length of the agents' trajectories using $l_T = v_i \cdot T$ with the time horizon $T$. By  cutting the map paths $\textbf{p}_i$ from the current agent position to the position $m=l_T$, we achieve a dynamic horizon. The second step of the trajectory distance model is then to calculate the minimal distance between both cut paths from the other agent and the ego agent. Hereby, the minimal distance is computed as in the path distance: the minimum value is taken over all point combinations. This is similar to assuming the agents to be at all predicted positions at the same time. Since it thus does not consider the time evolution of the trajectories, this resembles a worst-case approximation of the time models.  
The formula of the trajectory distance follows as
\begin{equation}
d_T(t)=\text{min}_{m_1 \in [0, v_1\cdot T],m_2 \in [0, v_2\cdot T]}\{||\textbf{p}_2(m_2) - \textbf{p}_1(m_1)||\} 
\end{equation}
The distance is transferred to a risk value by computing the inverse and obtaining $R_T(t)$. Hereby, a value for $T$ needs to be chosen for the time horizon.

Concerning the computation time, the trajectory distance uses $p^2$ operations like the path distance, with $p$ being this time the number of points in the paths' dynamic horizon. The time complexity follows to $\text{O}(p^2)$. Since one path is represented accurately with few straight line segments and the distance between two lines is easily computed analytically, the number of points $p$ can be set low so that the model is computationally cheap in its implementation.



\vspace{0.05cm}
\section{Experiments}
\label{sec:experiments}
In this section, we target to compare the performance and robustness of the considered risk models from this paper for the task of importance filtering. We will quantify which risk model works the best in filtering out most of the surrounding unimportant agents while keeping all important agents. The chosen driving situation is a complex intersection 
containing many agents driving on multiple paths with following and crossing behavior. Hereby, the dataset is synthetic and was created with an own simulator. For driving risk, there cannot be a ground truth available. In the case of filtering, we do not know the real set of important agents. Therefore, the filtered set of the model with the highest performance, the survival analysis, is considered in the experiments as the baseline. In this way, the risk models can be evaluated relative to each other.  

The section is divided into three parts. First, we will give two single examples of importance filtering with risk models to understand the differences between the risk models. Then, large-scale test results will be shown. We have applied the considered risk models on over five hundred variations of the chosen driving situation. 

\subsection{Single Examples}

\begin{figure}[t!]
  \centering
  \vspace{0.14cm}
  \hspace*{-0.3cm}
  \resizebox{0.49\linewidth}{!}{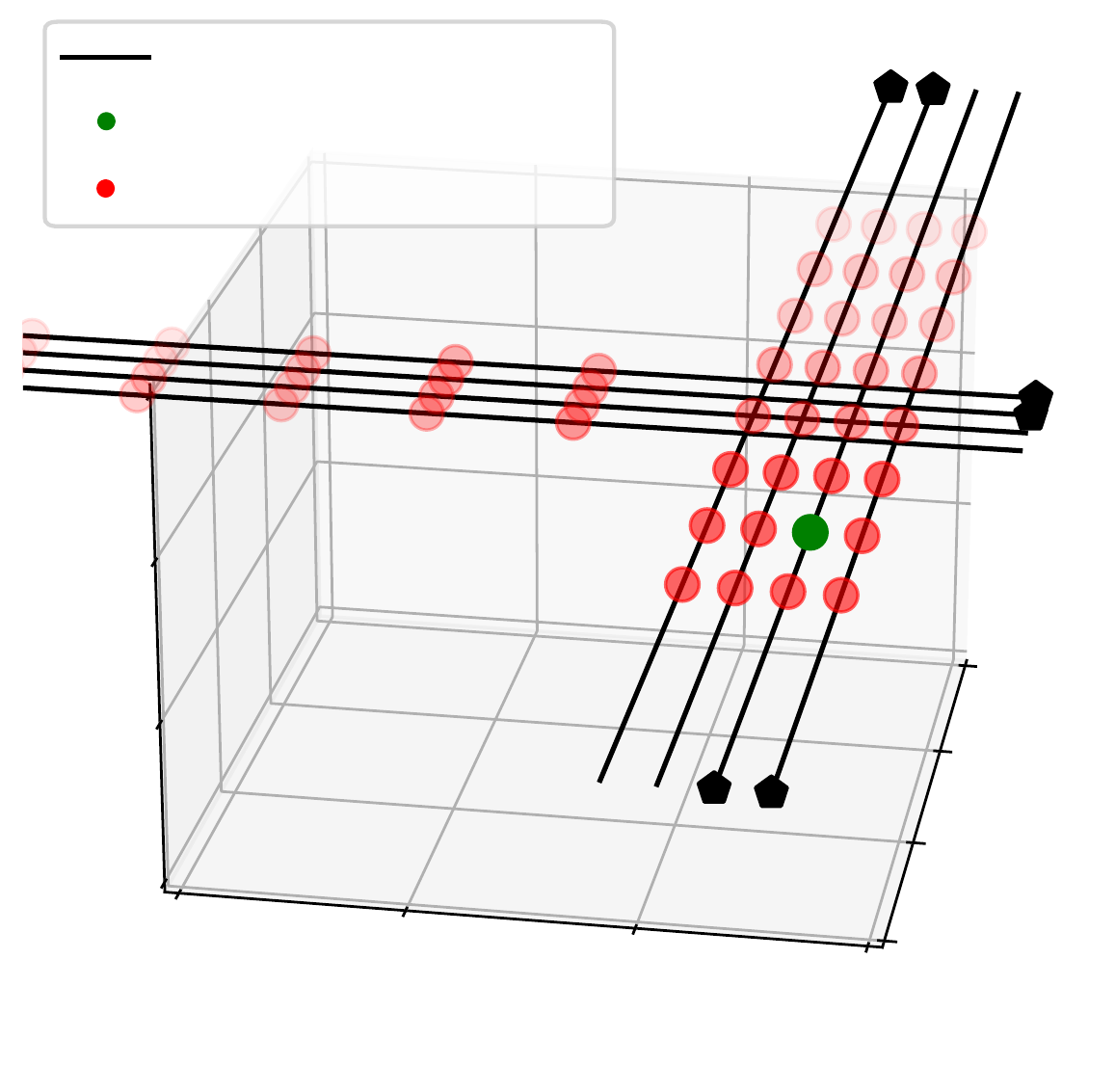}
  \resizebox{0.49\linewidth}{!}{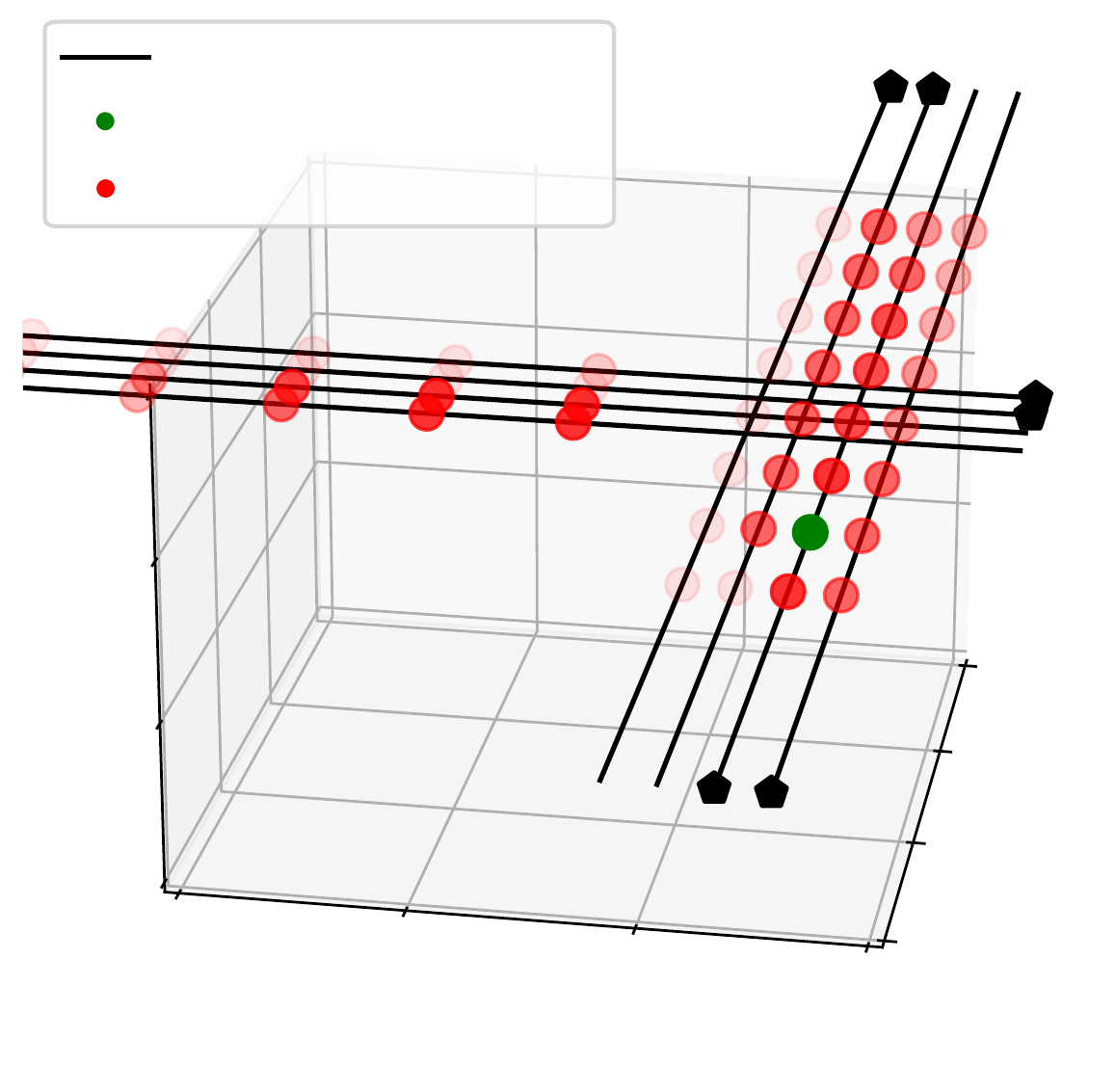}
  \vspace*{0.2cm}
  \caption[]{\hspace{0.07cm} Comparison of the simplest risk model, the current distance (left), and baseline model, the survival analysis (right), for an intersection driving situation with many agents. Agents with higher red color saturation are considered as more important. While the current distance considers closer agents as important, the survival analysis considers agents important that drive in the same direction or cross the ego agent's path in the future.
  }
  \vspace{-0.23cm}
  \label{fig:single_examples}
\end{figure}

To give a better understanding what the risk model comparison implies, two examples of importance from risks are shown in Fig. \ref{fig:single_examples}. In the figure, the ego agent is colored~in green and surrounding other agents in red. Every agent drives with the same velocity, which was chosen arbitrarily. On the left, the current distance model is applied for all $51$ other agents. The figure shows low/high importance by the low/high red color saturation. For the current distance model, the closer the agents are located the more important they are. In contrast, on the right, the application for the same driving situation with the survival analysis is given. 

The risk values are more fine-grained. The most important agents are other agents which drive on the same path, or which are crossing the ego agent's path. The second most important agents are agents which are right next to the ego path and are driving towards the ego agent. The third most important group of agents are agents on the neighboring lane driving in the same direction and the least important agents have already passed the ego agent or are further away on other paths.

These two models are at the extremes of the risk model overview from this paper. The current distance is the simplest model and has low performance and low computational cost. In contrast, the survival analysis has high performance and high computational cost. This model has already been shown to be superior compared to other risk models in the task of accident detection \cite{eggert2017}. In addition, the importance rating of agents was just shown to be also fine-grained and reasonable. 
We therefore assume the survival analysis as the baseline for importance filtering. 

\subsection{Large-scale Test}
In this second experiment, we will quantitatively compare the importance filtering results from all presented risk models in relation to the baseline of the survival analysis model. In particular, with a fixed risk threshold of $R_{S,{\text{th}}}\hspace{-0.05cm}=\hspace{-0.05cm}10^{-25}$ for the survival analysis, we check which other agents are filtered out and which are remaining and then we compare this set with the sets from each other risk model. In this way, the number of false negative errors and false positive errors can be computed. 
For each risk model, the risk threshold is then varied and we analyze how well the results compare to the survival analysis. 

\begin{figure}[t!]
  \centering
  \vspace{0.45cm}
  \resizebox{0.83\linewidth}{!}{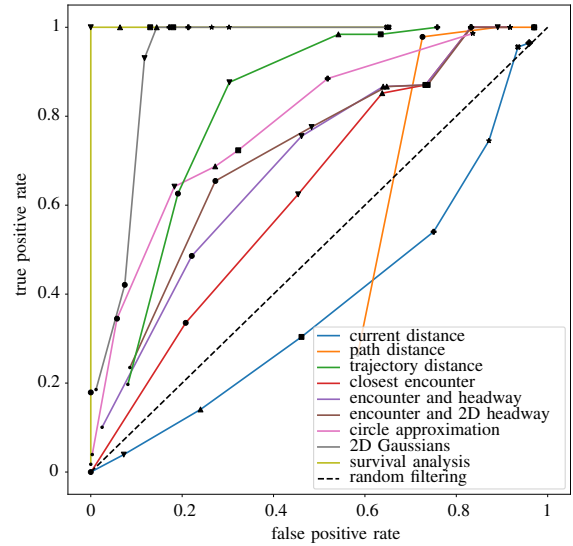}
  \vspace*{0.42cm}
  \caption[]{\hspace{0.07cm} Experiment results for the performance of all risk models considered in this paper. The plot shows the Receiver Operating Characteristics (ROC) curve for the task of importance filtering. Hereby, many variations of intersection situations were aggregated in the plot, with different agent numbers, distances and velocities. We consider the novel trajectory distance as the best compromise between performance and computational cost.}
  \vspace{-0.1cm}
  \label{fig:roc}
\end{figure}

For the experiment, we sample from our synthetic dataset around five hundred driving situations with a similar road layout as from the last single examples experiment. In the driving situations, the number of agents is overall in the range of $30$ to $100$, the distances between the agents between $\unit{5}{\text{m}}$ and $\unit{100}{\text{m}}$ and the driving velocities between  $\unit{0}{\text{m/s}}$ and $\unit{25}{\text{m/s}}$. A summary of the results from the experiments is given in a Receiver Operating Characteristics (ROC) curve depicted in Fig. \ref{fig:roc}. The plot shows for different risk thresholds of the risk models the True Positive Rate (TPR) over the False Positive Rate (FPR). 
Since we evaluated multiple driving situations, the ROC curve shows the average values. As we found out in Section \ref{sec:filtering_and_risk}, the indicator TPR is hereby more critical as it contains false negative errors, which are agents that are filtered out but which would actually be considered as important in the survival analysis, leading to unsafe plans. The ideal curve is close to the point with values of TPR~=~1 and FPR = 0. Analyzing the ROC curve allows us now to rate the models and show which models have bad and which models have good performance.

As the figure demonstrates: the current distance model is far away from the ideal point, see the blue curve. Since the situations are sampled with varying velocities, the current distance model cannot rate the importance correctly and has the worst performance. The path distance has either low TPR values or high FPR values (see orange curve), which is also not very close to the ideal point. Closest encounter without headway (red curve), with headway (purple curve) and with 2D headway (brown curve) are risk models with a better performance since they are closer to the top left area of the ROC plane. However, the robustness of these models is low. Their standard deviations in TPR are high with a value, for example, for the closest encounter with 2D headway of $\sigma_{\text{2D-TH}} = 0.14$. Here, the path distance has a better robustness with $\sigma_p = 0.04$. The circle approximation (see pink curve) can be rated slightly better than the closest encounter models, it has a higher performance and robustness. The 2D Gaussian method is visualized as a grey curve as well and this model is closest to the survival analysis in terms of performance. This is reasonable since the survival analysis uses the Gaussian overlap as a basis and weights the overlaps with a Poisson process function. As a baseline reference, the survival analysis is indicated in the figure as a yellow curve. 

Finally, the novel trajectory distance model is also depicted in Fig. \ref{fig:roc}. The model is colored in green and lies between the closest encounter with 2D headway and the 2D Gaussian method. In particular, its performance is better than from the closest encounter with 2D headway and it additionally possesses a better robustness with $\sigma_T$~$=$~$0.07$. Compared to the 2D Gaussian method, the model shows to have a worse performance but is computationally cheaper. Hereby, a value of $T =  \unit{12}{\text{s}}$ was chosen for the time horizon. 
This model therefore turned out to have a good performance, reasonable robustness and low computation time. Please note that the model can only be used for filtering tasks.


\section{Filtering Architecture}
\label{sec:filtering_architecture}
We have seen in the experiments that by using different risk thresholds, different ratios of False Positives (FP) and False Negatives (FN) are achieved in the importance filtering task with risk models, as illustrated in the Receiver Operating Characteristics curve. We have also seen that there are risk models with high performance, which means that the models induce a low number of FP and FN for one or several thresholds. These models also have higher computational costs. On the other side, the trajectory distance was shown to balance performance, robustness and computational costs~well. 

We consider the main novelty of this paper that we applied driving risk models for the task of importance filtering and show first evaluations of this kind. Based on the results, we however are also able to propose a novel filtering architecture that allows to reduce the complexity for behavior planning problems in complex urban driving. In the following, we therefore want to further outline how such a filtering architecture may look like. The filtering architecture will however not be experimentally evaluated.

\begin{figure}[t!]
  \centering

  \vspace*{0.24cm}
  \resizebox{0.66\linewidth}{!}{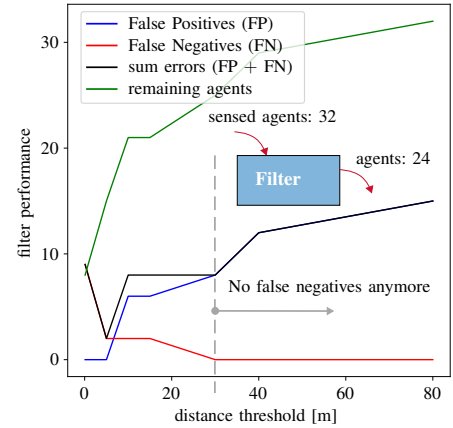}
  \vspace*{0.25cm}
  
  \caption[]{\hspace{0.07cm} In the filtering process, to avoid safety issues, it is crucial to reduce the number of false negatives. We recommend setting the threshold of the risk model so that there are few or no false negatives induced. In this way, multiple filters can be stacked to form a filtering architecture.
  \vspace*{-0.03cm}} 
  \label{fig:number_of_errors}
\end{figure}

Every model introduces specific misclassification errors. To avoid safety issues, it is crucial that the risk threshold is set in a way that more errors from FP and few errors from FN are induced. Fig. \ref{fig:number_of_errors} illustrates this risk thresholding dilemma exemplarily for the closest encounter with 2D headway. Note that for reasons of clarity, not the risk threshold is given in the figure but the distance threshold. For other risk models, the relationship looks qualitatively similar. Setting the distance threshold low creates FN (see red curve) but at some higher threshold there are no FN anymore, only some FP (see blue curve). Applying the risk model with this high distance threshold, which means, in turn, a low risk threshold, allows us to achieve a safe importance filtering. We could reduce the sensed agents from a number of, for example, 32 to 24 agents, with only few false positives.  With this relationship in mind, we can actually stack different filters to further reduce computation time. The strategy is to use simpler models first with a correct threshold and then step-by-step filter with more complex risk models, inducing nearly no FN in the process of the overall system. 

\subsection{Conceptual Architecture Proposal}
\begin{figure}[t!]
  \centering
  \vspace*{0.4cm}
  \resizebox{\linewidth}{!}{\import{./images/}{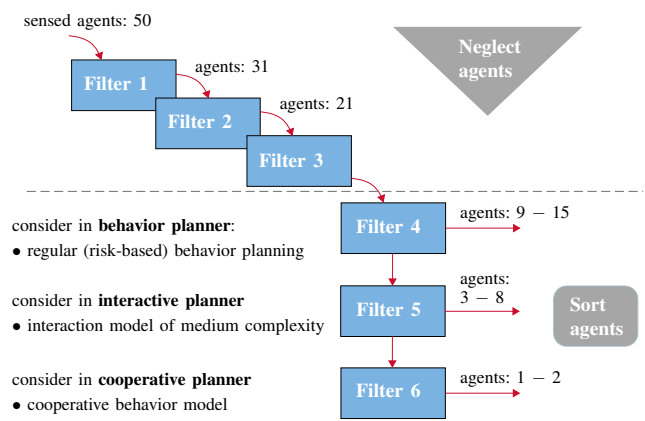}}
  
  \vspace*{0.15cm}
  \caption[]{\hspace{0.07cm} The filtering architecture consists of two parts: first, incrementally filtering out agents with multiple filters and second, sorting agents based on their importance. For the second part, planners of increasing complexity can be used to be able to handle complex driving situations with higher efficiency and robustness. The figure gives an example of how a set of fifty sensed agents would be handled in the architecture.} 
  \label{fig:filtering_architecture}
\end{figure}

The proposed architecture for filtering is comprised of two parts, one part where less important surrounding agents are subsequently filtered out with different risk models, starting with computationally cheaper models first to save time. The recommendation based on the experiments in this paper is to first filter agents out using the path distance model because its robustness is high, then with the trajectory distance with its reasonable performance versus computational cost and in the end with the 2D Gaussian method which is having an even better performance. In a second part of the architecture, the remaining agents are sorted by clustering agents in groups based on their importance which are handled differently in the upcoming planning system. The sorting must be done with a more complex model than in the previous step-by-step neglecting of agents. In this case, it should be performed with the most accurate survival analysis using different risk thresholds. Fig. \ref{fig:filtering_architecture} gives an overview of the overall filtering architecture.


This filtering architecture has several advantages. On the one hand side, neglecting agents will reduce the computational costs of the evaluation in the final planner. While the risk models need to be applied only once in the filter, the risk model is used in an optimization with many evaluations in the planning system. In addition, the used risk models~are computationally cheaper in the filtering step
and in the final planner the most complex model has to be used, which further reduces the computation time.
On the other hand side, sorting the agents afterwards allows to plan with different planners which consider the agents differently to additionally improve the overall robustness of the system.

The figure gives an example how a set of 50 sensed agents would be handled in the filtering architecture. After applying the first filter, only 31 agents are considered important, after the second filter 21 agents and after the third filter 15 agents. Then, the sorting step is applied. In the sorting step, on the highest layer, around 9\hspace{0.06cm}-\hspace{0.06cm}15 agents may be considered in the behavior planner. The behavior planner solely applies a regular risk-based planning method with lower computational cost. By sorting out more important agents in this set, we can retrieve 3\hspace{0.06cm}-\hspace{0.06cm}8 agents which are also used in an interactive planner on the middle layer. This planner could, for example, use an interaction model with medium complexity to solve more advanced driving situations. Finally, after applying the last filter, on the lowest layer only 1\hspace{0.06cm}-\hspace{0.05cm}2 agents are considered in a complex cooperative planner which includes a complete cooperative model. 

Applying multiple planners allows to save more computation time in crowded, complex situations. Furthermore, this adds safety and robustness. The planners can inform each other for agents that are, for example, in both layers and one can switch between the planners if one planner cannot find a solution. 

\section{Conclusion and Outlook}
\label{sec:conclusion}
In conclusion, this paper targeted to introduce the research field of importance filtering with driving risk models. After presenting the challenges of importance filtering and giving an overview of risk models, we compared the risk models on the basis of their performance and robustness for a dataset of driving situations. The results showed that the novel trajectory distance of this paper balances performance, robustness and computation time well. In addition, the paper then also presented a novel conceptual system architecture for filtering in crowded environments. For the system architecture, we finally recommended which risk models to use in each filtering and sorting step. 

We consider the contributions of this paper to be therefore three-fold: 1) a novel application of and comparison of risk models for filtering, 2) a novel compact trajectory distance model and 3) a  first conceptual proposal of a novel filtering architecture.

Filtering with driving risk models enables to scale complex behavior planners beyond smaller interaction-intensive driving situations with few surrounding sensed agents. We envision complex behavior planners using the filtering architecture to tackle large driving situations with more than fifty surrounding agents. The risk models can be naturally used for other physical embodiments besides self-driving cars, such as for vehicles in marine, robotics, aviation and space. The interactions are similar and contain time-based motion with uncertainty. For mobile robots moving in pedestrian crowds, this can be especially helpful because they face many other agents during use.

For future work, we want to apply the importance filtering on real driving data with complex situations, for example, on the Waymo Open dataset which can be found in \cite{waymo_open_dataset}. In particular, we target to evaluate the advantages of the filtering architecture in combination with a planning system to show that accident rates are not increased with the filtering, while the overall computational costs are reduced.  

\addtolength{\textheight}{-12cm}   

\bibliographystyle{IEEEtran}
\bibliography{bib}
\end{document}